\documentclass{article}
\PassOptionsToPackage{numbers, compress, sort}{natbib}

\usepackage[preprint]{neurips_2023}

\usepackage[utf8]{inputenc} 
\usepackage[T1]{fontenc}    
\usepackage[pagebackref=true,breaklinks=true,letterpaper=true,colorlinks,bookmarks=false]{hyperref}       
\usepackage{url}            
\usepackage{booktabs}       
\usepackage{amsfonts}       
\usepackage{nicefrac}       
\usepackage{microtype}      
\usepackage{xcolor}         
\usepackage[font=small]{caption}

\newcommand{\eqn}[1]{Equation~\ref{#1}}

\newcommand{\ignore}[1]{}

\renewcommand*{\thefootnote}{\fnsymbol{footnote}}

\DeclareMathAlphabet{\mathbfit}{OML}{cmm}{b}{it}

\makeatletter
\DeclareRobustCommand\onedot{\futurelet\@let@token\@onedot}
\def\@onedot{\ifx\@let@token.\else.\null\fi\xspace}

\def\eg{e.g\onedot}

\def\method{FastComposer\xspace}

\newcommand{\myparagraph}[1]{\vspace{0pt}\paragraph{#1}}

\usepackage{hyperref}

\usepackage{times}
\usepackage{latexsym}
\usepackage{amsmath}

\usepackage[T1]{fontenc}
\usepackage[utf8]{inputenc}

\usepackage{microtype}
\usepackage{bbding}
\usepackage{color,xcolor}
\usepackage{epsfig}
\usepackage{graphicx}

\usepackage{adjustbox}
\usepackage{array}
\usepackage{booktabs}
\usepackage{colortbl}
\usepackage{wrapfig}
\usepackage{hhline}
\usepackage{multirow}
\usepackage{subcaption} 

\usepackage{amsmath,amsfonts,amssymb}
\usepackage{bm}
\usepackage{nicefrac}
\usepackage{microtype}
\usepackage{mathtools}

\usepackage{changepage}
\usepackage{extramarks}
\usepackage{fancyhdr}
\usepackage{lastpage}
\usepackage{setspace}
\usepackage{soul}
\usepackage{xspace}

\usepackage{url}

\usepackage{enumerate}
\usepackage{enumitem}  
\usepackage{titlesec}

\usepackage{makecell}

\usepackage{pifont} 

\usepackage{amsthm}
\usepackage{float}

\title{\method: Tuning-Free Multi-Subject Image Generation with Localized Attention}

\author{
Guangxuan Xiao$^{*}$
\space
Tianwei Yin$^{*}$
\space
William T. Freeman
\space
Fr\'edo Durand
\space
Song Han \\
Massachusetts Institute of Technology 
\\
\url{https://fastcomposer.mit.edu}
}

\begin{document}

\maketitle
\let\thefootnote\relax\footnotetext{$^{*}$ Equal Contribution.}
\let\thefootnote\relax\footnotetext{Correspondence to: Guangxuan Xiao <xgx@mit.edu>, Tianwei Yin <tianweiy@mit.edu>.}

\begin{abstract}
Diffusion models excel at text-to-image generation, especially in subject-driven generation for personalized images. 
However, existing methods are inefficient due to the subject-specific fine-tuning, which is computationally intensive and hampers efficient deployment.
Moreover, existing methods struggle with multi-subject generation as they often blend identity among subjects.
We present \method which enables efficient, personalized, multi-subject text-to-image generation without fine-tuning.
\method uses subject embeddings extracted by an image encoder to augment the generic text conditioning in diffusion models, enabling personalized image generation based on subject images and textual instructions \textit{with only forward passes}. 
To address the identity blending problem in the multi-subject generation, \method proposes \emph{cross-attention localization} supervision during training, enforcing the attention of reference subjects localized to the correct regions in the target images. 
Naively conditioning on subject embeddings results in subject overfitting. \method proposes \emph{delayed subject conditioning} in the denoising step to maintain both identity and editability in subject-driven image generation. 
\method generates images of multiple unseen individuals with different styles, actions, and contexts.  
It achieves 300$\times$-2500$\times$ speedup compared to fine-tuning-based methods and requires zero extra storage for new subjects. \method paves the way for efficient, personalized, and high-quality multi-subject image creation.
Code, model, and datasets will be released for reproduction.
\end{abstract}

\section{Introduction}
\begin{figure*}[t]
    \centering
    \includegraphics[width=\linewidth]{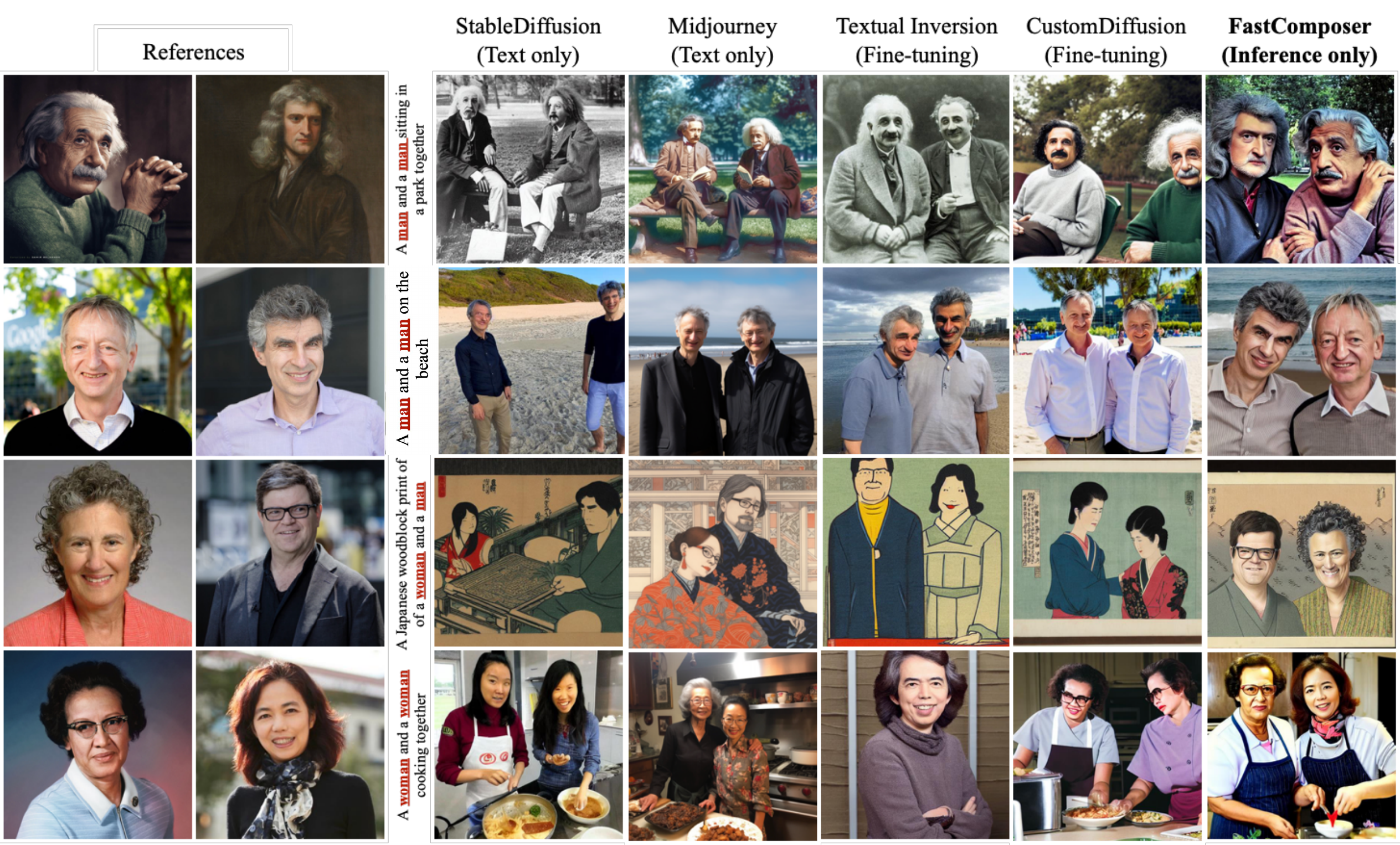}
    \caption{Comparison with baselines for multi-subject image generation.
    We use scientists' names in the text prompt for text-only methods (SD, MJ).
    Text-only methods only perform well when subjects are present in the training dataset but struggle to maintain the identity otherwise.
    Fine-tuning-based methods blend the identity of different persons (TI rows 1 and 2, CD rows 1, 2, 4), deviate from the text instruction and only generate a single subject (TI row 4), or generate images that do not resemble any specific reference (CD row 3).}
    \vspace{-0.5em}
    \label{fig:multi_subject}
\end{figure*}
\begin{figure*}[t]
    \centering
    \includegraphics[width=\linewidth]{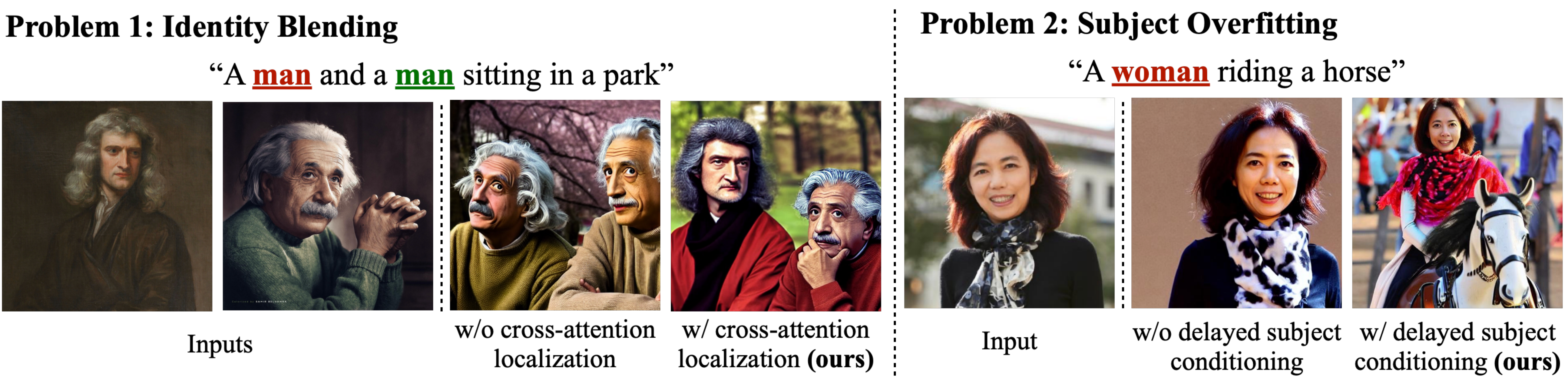}
    \caption{
    Two challenges faced by existing subject-driven image generation methods. 
    Firstly, current methods blend the distinct characteristics of different subjects~(\textbf{identity blending}), shown by the right figure where Newton resembles Einstein.
     \textit{Cross-attention localization}~(Sec~\ref{sec:crossattn}) solves this problem.
     Secondly, they suffer from \textbf{subject overfitting}, where they overfit the input image and ignore the text instruction.
    \textit{Delayed subject conditioning} (Sec~\ref{sec:guidance}) addresses this issue. 
}  
    \vspace{-1em}
    \label{fig:problem}
\end{figure*}

Recent advancements in text-to-image generation~\cite{ramesh2021zero, chang2023muse, kang2023scaling, ding2021cogview}, particularly diffusion models~\cite{ho2020denoising, song2020score, rombach2022high, ramesh2022hierarchical, sohl2015deep}, have opened new frontiers in content creation. 
Subject-driven text-to-image generation permits the personalization to new individuals given a few sample images~\cite{ruiz2022dreambooth, casanova2021instance, nitzan2022mystyle, gal2022image, kumari2022multi}, allowing the generation of images featuring specific subjects in novel scenes, styles, and actions. 
However, current subject-driven text-to-image generation methods suffer from two key limitations: cost of personalization and identity blending for multiple subjects. 
Personalization is costly because they often need model fine-tuning for each new subject for best fidelity. 
The computational overhead and high hardware demands introduced by model tuning, largely due to the memory consumption\cite{chen2016training} and computations of backpropagation, constrain the applicability of these models across various platforms.
Furthermore, existing techniques struggle with multi-subject generation (Figure~\ref{fig:multi_subject}) because of the ``identity blending'' issue (Figure~\ref{fig:problem} left), in which the model combines the distinct characteristics of different subjects (subject A looks like subject B and vice versa).

We propose \method, a tuning-free, personalized multi-subject text-to-image generation method. 
Our key idea is to replace the generic word tokens, such as "person", by an embedding that captures an individual's unique identity in the text conditioning. 
 We use a vision encoder to derive this identity embedding from a referenced image, and then augment the generic text tokens with features from this identity embedding.
 This enables image generation based on subject-augmented conditioning.
Our design allows the generation of images featuring specified subjects with only forward passes and can be further integrated with model compression techniques~\cite{xiao2022smoothquant, bolya2022token, han2015deep} to boost deployment efficiency.

To tackle the multi-subject identity blending issue, we identify unregulated cross-attention as the primary reason (Figure~\ref{fig:crossattn}). 
When the text includes two "person" tokens, each token's attention map attends to both person in the image rather than linking each token to a distinct person in the image. 
To address this, we propose supervising cross-attention maps of subjects with segmentation masks during training (i.e., \emph{cross-attention localization}), using standard segmentation tools~\cite{cheng2022masked}. This supervision explicitly guides the model to map subject features to distinct and non-overlapping regions of the image, thereby facilitating the generation of high-quality multi-subject images (Figure~\ref{fig:problem} left).
We note that segmentation and cross-attention localization is only required during the training phase. 

Naively applying subject-augmented conditioning leads to subject overfitting (Figure~\ref{fig:problem} right), restricting the user's ability to edit subjects based on textual directives. To address this, we introduce \emph{delayed subject conditioning}, preserving the subject's identity while following text instructions. It employs text-only conditioning in the early denoising stage to generate the image layout, followed by subject-augmented conditioning in the remaining denoising steps to refine the subject appearance. This simple technique effectively preserves subject identity without sacrificing editability (Figure~\ref{fig:timesteps}).

For the first time, \method enables inference-only generation of multiple-subject images across diverse scenarios~(Figure~\ref{fig:multi_subject}). 
\method achieves 300$\times$-2500$\times$ speedup and 2.8$\times$-6.7$\times$ memory saving compared to fine-tuning-based methods, requiring zero extra storage for new subjects. \method paves the way for low-cost, personalized, and versatile text-to-image generation. 

\section{Related Work}
\textbf{Subject-Driven Image Generation} aims to render a particular subject unseen at the initial training stage. Given a limited number of example images of the subject, it seeks to synthesize novel renditions in diverse contexts.
DreamBooth~\cite{ruiz2022dreambooth}, textual-inversion~\cite{gal2022image}, and custom-diffusion~\cite{kumari2022multi} use optimization-based methods to embed subjects into diffusion models.
This is achieved by either fine-tuning the model weights~\cite{ruiz2022dreambooth, kumari2022multi} or inverting the subject image into a text token that encodes the subject identity~\cite{gal2022image}. 
Recently, tuning-encoder~\cite{roich2022pivotal} reduces the total number of fine-tuning steps by first generating an inverted set latent code using a pre-trained encoder and then refines these codes through several finetuning steps to better preserve subject identities.
However, all these tuning-based methods~\cite{gal2023designing, kumari2022multi, gal2022image, ruiz2022dreambooth} require resource-intensive backpropagation, and the hardware must be capable of fine-tuning the model, which is neither feasible on edge devices such as smartphones, nor scalable for cloud-based applications.
In contrast, our new \method amortizes the costly subject tuning during the training phase, enabling instantaneous personalization of multiple subjects using simple feedforward methods at test time.

A number of concurrent works have explored tuning-free methods. 
X\&Fuse~\cite{kirstain2023x} concatenates the reference image with the noisy latent for image conditioning. 
ELITE~\cite{wei2023elite} and InstantBooth~\cite{shi2023instantbooth} use global and local mapping networks to project reference images into word embeddings and inject reference image patch features into cross-attention layers to enhance local details.
Despite impressive results for single-object customization, their architecture design restricts their applicability to multiple subject settings, as they rely on global interactions between the generated image and reference input image.
UMM-Diffusion~\cite{ma2023unified} shares a similar architecture to ours.
However, it faces identity blending challenges when extended to multiple subjects~\cite{ma2023unified}.
In comparison, our method supports multi-subject composition via a \textit{cross-attention localization} supervision mechanism~(Sec \ref{sec:crossattn}). 

\paragraph{Multi-Subject Image Generation.}
Custom-Diffusion~\cite{kumari2022multi} enables multi-concepts composition by jointly fine-tuning the diffusion model for multiple concepts. 
However, it typically handles concepts with clear semantic distinctions, such as animals and their related accessories or backgrounds. 
The method encounters challenges when dealing with subjects within similar categories, often generating the same person twice when composing two different individuals ~(Figure \ref{fig:multi_subject}).
SpaText~\cite{avrahami2022spatext}, and Collage Diffusion~\cite{sarukkai2023collage} enable multi-object composition through a layout to image generation process.
A user-provided segmentation mask determines the final layouts, which are then transformed into high-resolution images using a diffusion model. 
Nevertheless, these techniques either compose generic objects without customization~\cite{avrahami2022spatext} or demand the costly textual-inversion process to encode instance-specific details~\cite{sarukkai2023collage}.
Furthermore, these techniques require a user-provided segmentation map.
In contrast, \method generates personalized, multi-subject images in an inference-only manner and automatically derives plausible layouts from text prompts.

\begin{figure*}
    \centering
    \includegraphics[width=\linewidth]{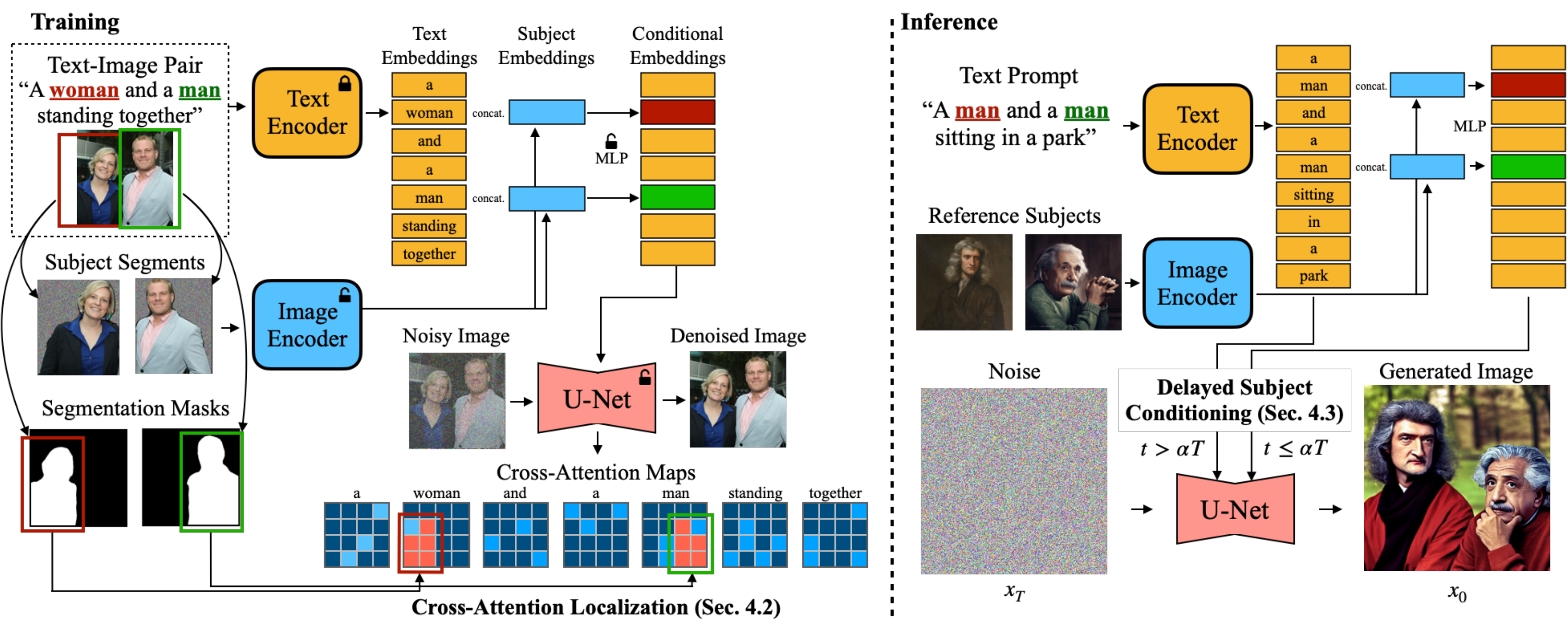}
    \caption{Training and inference pipeline of \method. Given a text description and images of multiple subjects, \method uses an image encoder to extract the features of the subjects and augments the corresponding text tokens. 
    The diffusion model is trained to generate multi-subject images with augmented conditioning.
    We use \textit{cross-attention localization}~(Sec. \ref{sec:crossattn}) to boost multi-subject generation quality, and \textit{delayed subject conditioning} to avoid subject overfitting~(Sec. \ref{sec:guidance}). }
    \vspace{-1em}
    \label{fig:training_inference}
\end{figure*}
\section{Preliminaries}

\myparagraph{Stable Diffusion.}
We use the state-of-the-art StableDiffusion (SD) model as our backbone network.
The SD model consists of three components: the variational autoencoder (VAE), U-Net, and a text encoder.
The VAE encoder $\mathcal{E}$ compresses the image $x$ to a smaller latent representation $z$, which is subsequently  perturbed by Gaussian noise $\varepsilon$ in the forward diffusion process.
The U-Net, parameterized by $\theta$, denoises the noisy latent representation by predicting the noise.
This denoising process can be conditioned on text prompts through the cross-attention mechanism, while the text encoder $\psi$ maps the text prompts $\mathcal{P}$ to conditional embeddings $\psi(\mathcal{P})$.
During training, the network is optimized to minimize the loss function given by the equation below:
\begin{equation}
    \label{eq:sd_loss}
    \mathcal{L}_{\text{noise}} =
    \mathbb{E}_{z\sim\mathcal{E}(x),\mathcal{P},\varepsilon\sim\mathcal{N}(0,1),t}
    \left [ || \varepsilon - \varepsilon_\theta(z_t, t, \psi(\mathcal{P})) ||_2^2 \right ],
\end{equation}
where $z_t$ is the latent code at time step $t$.
At inference time, a random noise $z_T$ is sampled from $\mathcal{N}(0,1)$ and iteratively denoised by the U-Net to the initial latent representation $z_0$. Finally, the VAE decoder $\mathcal{D}$ generates the final image by mapping the latent codes back to pixel space $\hat{x} = \mathcal{D}(z_0)$.

\myparagraph{Text-Conditioning via Cross-Attention Mechanism.}
\label{sec:prelim_crossattn}
In the SD model, the U-Net employs a cross-attention mechanism to denoise the latent code conditioned on text prompts. For simplicity, we use the single-head attention mechanism in our discussion. Let $\mathcal{P}$ represent the text prompts and $\psi$ denote the text encoder, which is typically a pre-trained CLIP text encoder. The encoder converts $\mathcal{P}$ into a list of $d$-dimensional embeddings, $\psi(\mathcal{P}) = c \in \mathbb{R}^{n\times d}$.
The cross-attention layer accepts the spatial latent code $z \in \mathbb{R}^{(h \times w) \times f}$ and the text embeddings $c$ as inputs. It then projects the latent code and text embeddings into Query, Key, and Value matrices: $Q = W^q z$, $K = W^k c$, and $V = W^v c$. Here, $W^q \in \mathbb{R}^{f \times d'}, W^k, W^v \in \mathbb{R}^{d \times d'}$ represent the weight matrices of the three linear layers, and $d'$ is the dimension of Query, Key, and Value embeddings.
The cross-attention layer then computes the attention scores $A = \text{Softmax}(\frac{QK^T}{\sqrt{d'}}) \in [0,1]^{(h \times w) \times n}$,
and takes a weighted sum over the Value matrix to obtain the cross-attention output $z_\text{attn} = AV \in \mathbb{R}^{(h \times w) \times d'}$.
Intuitively, the cross-attention mechanism ``scatters'' textual information to the 2D latent code space, and $A[i, j, k]$ represents the amount of information flow from the $k$-th text token to the $(i, j)$ latent pixel. Our method is  based on this semantic interpretation of the cross-attention map, and we will discuss it in detail in Section~\ref{sec:crossattn}.
\section{\method}
\label{sec:method}
\subsection{Tuning-Free Subject-Driven Image Generation with an Image Encoder}
\myparagraph{Augmenting Text Representation with Subject Embedding.} 
To achieve tuning-free subject-driven image generation, we propose to augment text prompts with  visual features extracted from reference subject images. 
Given a text prompt $\mathcal{P} = \{w_1, w_2, \dots w_n\}$, a list of reference subject images $\mathcal{S} = \{s_1, s_2, \dots s_m\}$, and an index list indicating which subject corresponds to which word in the text prompt $\mathcal{I} = \{i_1, i_2, \dots i_m\}, i_j \in {1, 2, \dots, n}$, we first encode the text prompt $\mathcal{P}$ and reference subjects $\mathcal{S}$ into embeddings using the pre-trained CLIP text and image encoders $\psi$ and $\phi$, respectively. Next, we employ a multilayer perceptron (MLP) to augment the text embeddings with visual features extracted from the reference subjects. We concatenate the word embeddings with the visual features and feed the resulting augmented embeddings into the MLP. This process yields the final conditioning embeddings $c' \in \mathbb{R}^{n\times d}$, defined as follows:
\begin{equation}
    c'_{i} = \begin{cases}
        \psi(\mathcal{P})_i,                            & i \notin \mathcal{I}   \\
        \text{MLP}(\psi(\mathcal{P})_i || \phi(s_{j})), & i = i_j\in \mathcal{I}
    \end{cases}
    \label{equ:augment}
\end{equation}
Figure~\ref{fig:training_inference} gives a concrete example of our augmentation approach. 

\myparagraph{Subject-Driven Image Generation Training.}
To enable inference-only subject-driven image generation, we train the image encoder, the MLP module, and the U-Net with the denoising loss~(Figure~\ref{fig:training_inference}).
We create a subject-augmented image-text paired dataset to train our model, where noun phrases from image captions are paired with subject segments appearing in the target images.
We initially use a dependency parsing model to chunk all noun phrases (e.g., ``a woman'') in image captions and a panoptic segmentation model to segment all subjects present in the image.
We then pair these subject segments with corresponding noun phrases in the captions with a greedy matching algorithm based on text and image similarity~\cite{radford2021learning, reimers2019sentence}.
The process of constructing the subject-augmented image-text dataset is detailed in Sec.~\ref{sec:dataset_construction}.
In the training phase, we employ subject-augmented conditioning, as outlined in \eqn{equ:augment}, to denoise the perturbed target image.
We also mask the subjects' backgrounds with random noise before encoding, preventing the overfitting of the subjects' backgrounds.
Consequently, \method can directly use natural subject images during inference without explicit background segmentation.

\subsection{Localizing Cross-Attention Maps with Subject Segmentation Masks}
\label{sec:crossattn}
\begin{figure*}[t]
    \centering
    \includegraphics[width=\linewidth]{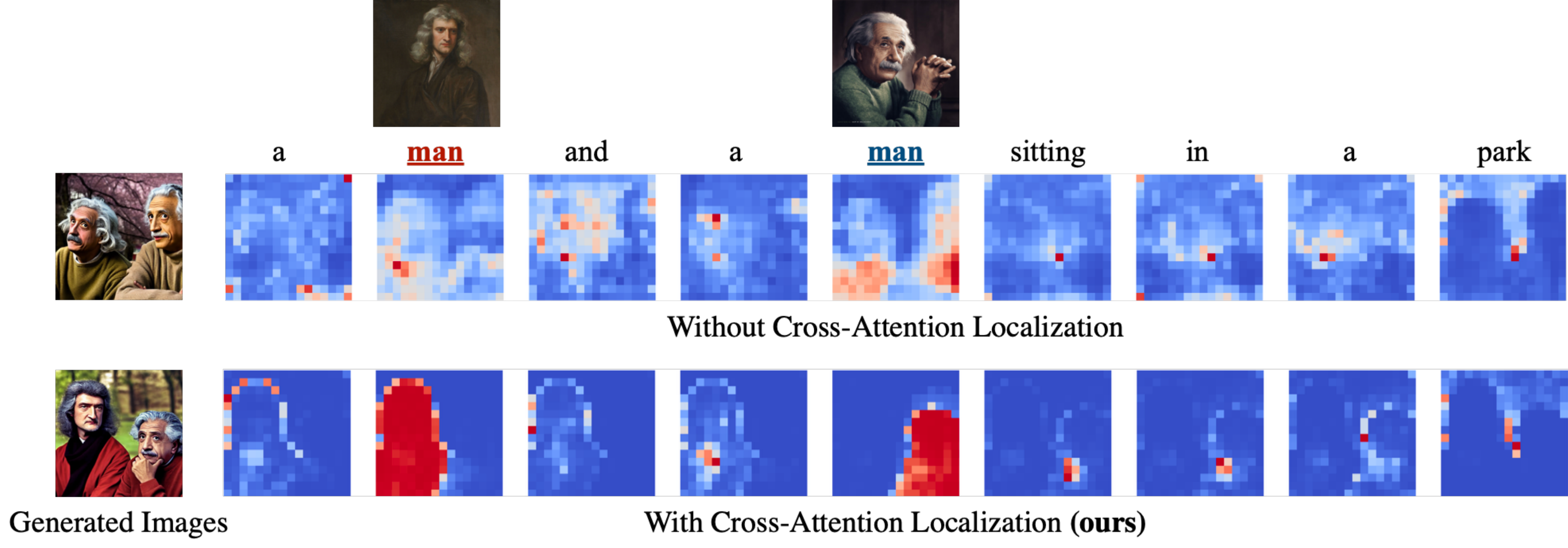}
    \caption{In the absence of cross-attention regularization (top), the diffusion model attends to multiple subjects' input tokens and merge their identity. By applying cross-attention regularization (bottom), the diffusion model learns to focus on only one reference token while generating a subject. This ensures that the features of multiple subjects in the generated image are more separated.}
    \vspace{-1em}
    \label{fig:crossattn}
\end{figure*}

We  observe that traditional cross-attention maps tend to attend to all subjects at the same time, which leads to identity blending in multi-subject image generation~(Figure~\ref{fig:crossattn} top). We propose to localize cross-attention maps with subject segmentation masks during training to solve this issue. 
\myparagraph{Understanding the Identity Blending in Diffusion Models.} Prior research~\cite{hertz2022prompt} shows that the cross-attention mechanism within diffusion models governs the layout of generated images. The scores in cross-attention maps represent ``the amount of information flows from a text token to a latent pixel.''
We hypothesize that identity blending arises from the unrestricted cross-attention mechanism, as a single latent pixel can attend to all text tokens.
If one subject's region attends to multiple reference subjects, identity blending will occur.
In Figure~\ref{fig:crossattn}, we confirm our hypothesis by visualizing the average cross-attention map within the U-Net of the diffusion model. The unregularized model often has two reference subject tokens influencing the same generated person at the same time, causing a mix of features from both subjects.
We argue that proper cross-attention maps should resemble an instance segmentation of the target image, clearly separating the features related to different subjects. To achieve this, we add a regularization term to the subject cross-attention maps during training to encourage focusing on specific instance areas. 
Segmentation maps and cross-attention regularization are only used during training, not at test time. 

\myparagraph{Localizing Cross-Attention with Segmentation Masks.}
As discussed in Section~\ref{sec:prelim_crossattn}, a cross-attention map $A \in [0,1]^{(h \times w) \times n}$ connects latent pixels to conditional embeddings at each layer, where $A[i, j, k]$ denotes the information flow from the $k$-th conditional token to the $(i, j)$ latent pixel. Ideally, the subject token's attention map should focus solely on the subject region rather than spreading throughout the entire image, preventing identity blending among subjects. To accomplish this, we propose localizing the cross-attention map using the reference subject's segmentation mask. Let $\mathcal{M} = \{M_1, M_2, \dots M_m\}$ represent the reference subjects' segmentation masks, $\mathcal{I} = \{i_1, i_2, \dots i_m\}$ be the index list indicating which subject corresponds to each word in the text prompt, and $A_{i} = A[:,:,i] \in [0,1]^{(h \times w)}$ be the cross-attention map of the $i$-th subject token.
We supervise the cross-attention map $A_{i_j}$ to be close to the segmentation mask $m_j$ of the $j$-th subject token, i.e., $A_{i_j} \approx m_j$. We employ a balanced L1 loss to minimize the distance between the cross-attention map and the segmentation mask:
\begin{equation}
    \mathcal{L}_{\text{loc}} = \frac{1}{m}\sum_{j=1}^m (\text{mean}(A_{i_j}[\bar{m}_j]) - \text{mean}(A_{i_j}[m_j])).
\end{equation}
The final training objective of \method is given by:
\begin{equation}
    \mathcal{L} = \mathcal{L}_{\text{noise}} + \lambda \mathcal{L}_{\text{loc}},
\end{equation}
using a localization loss ratio controlled by hyperparameter $\lambda = 0.001$. Motivated by~\cite{hertz2022prompt, chefer2023attend}, we apply the localization loss to the downsampled cross-attention maps, i.e., the middle 5 blocks of the U-Net, which are known to contain more semantic information.
As illustrated in Figure~\ref{fig:crossattn}, our localization technique enables the model to precisely allocate attention to reference subjects at test time, which prevents identity blending between subjects.

\begin{figure*}[t]
    \centering
    \includegraphics[width=\linewidth]{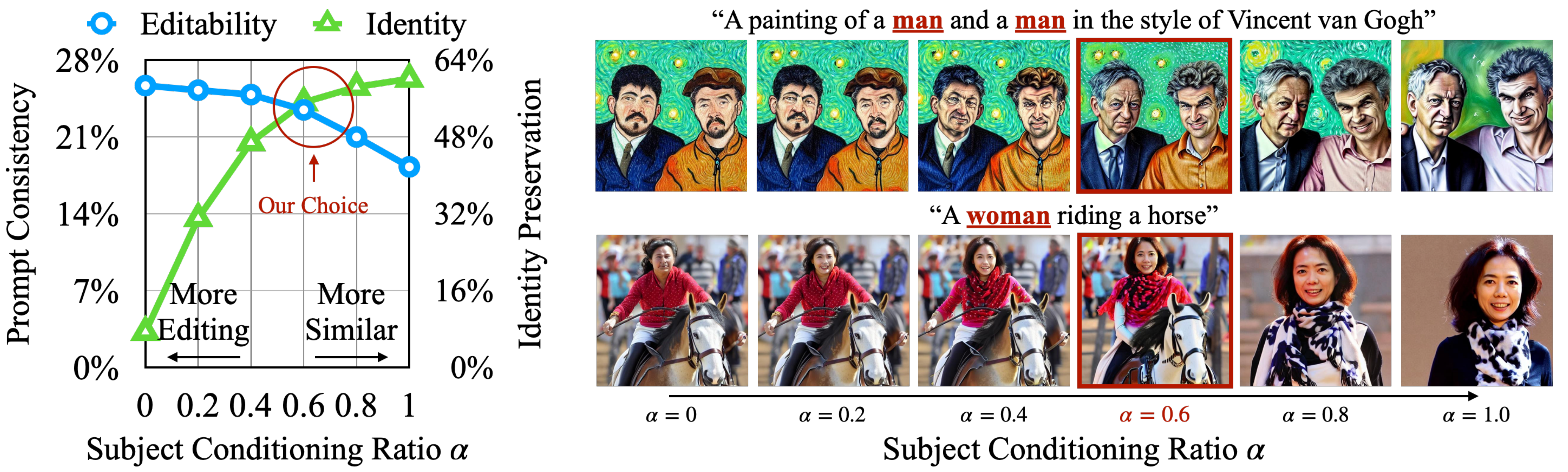}
    \caption{Effects of using different ratios of timesteps for subject conditioning. A ratio between 0.6 to 0.8 yields good results and achieve a balance between prompt consistency and identity preservation.}
    \vspace{-1em}
    \label{fig:timesteps}
\end{figure*}
\subsection{Delayed Subject Conditioning in Iterative Denoising}
\label{sec:guidance}
During inference, using the augmented text representation directly often leads to images that closely resemble the subjects while ignoring the textual directives.
This occurs because the image layout forms at the early phases of the denoising process, and premature augmentation from the reference image causes the resulting image to stray from the text instructions.
Prior methods~\cite{gal2023designing, roich2022pivotal} mitigate this issue by generating an initial latent code and refining it through iterative model finetuning.
However, this process is resource-intensive and needs high-end devices for model fine-tuning.
Inspired by Style Mixing~\cite{karras2019style}, we propose a simple \emph{delayed subject conditioning}, which allows for inference-only subject conditioning while striking a balance between identity preservation and editability.

Specifically, we perform image augmentation only after the layout has been created using a text-only prompt.
In this framework, our time-dependent noise prediction model can be represented as:
\begin{equation}
    \epsilon_t =
    \begin{cases}
        \epsilon_\theta(z_t, t, c)  & \text{if } t > \alpha T, \\
        \epsilon_\theta(z_t, t, c') & \text{otherwise}
    \end{cases}
\end{equation}
Here, $c$ denotes the original text embedding and $c'$ denotes text embedding augmented with the input image embedding. $\alpha$ is a hyperparameter indicating the ratio of subject conditioning.
We ablate the effect of using different $\alpha$ in Figure~\ref{fig:timesteps}.
Empirically, $\alpha \in [0.6, 0.8]$ yields good results that balance prompt consistency and identity preservation, though it can be easily tuned for specific instances.

\begin{figure*}[t]
    \centering
    \includegraphics[width=\linewidth]{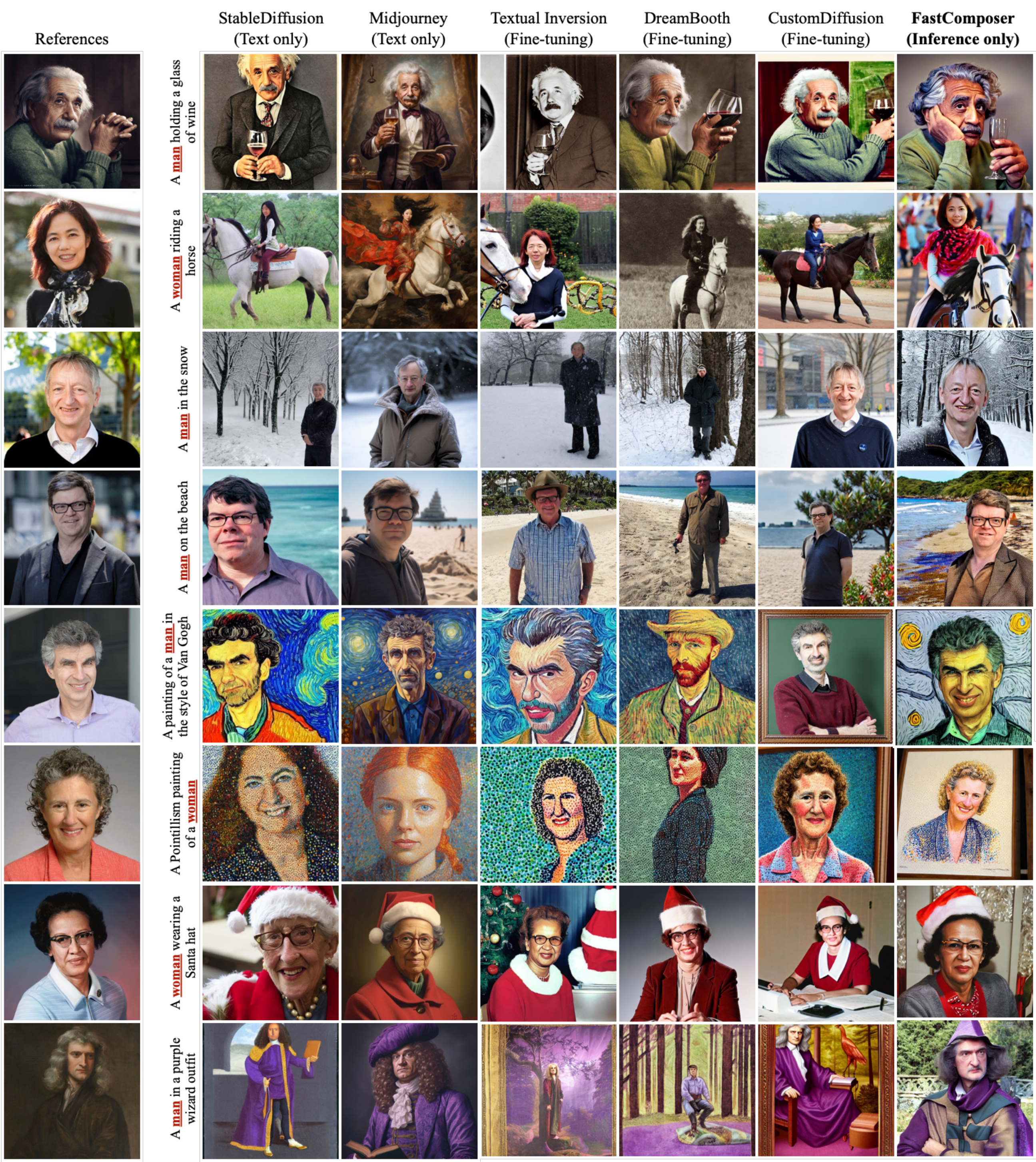}
    \caption{Comparison of different methods on single subject image generation. For text-only methods (i.e., StableDiffusion and Midjourney), we use scientists’ names in the text prompt.}
    \vspace{-1em}
    \label{fig:single_subject}
\end{figure*}
\section{Experiments}
\subsection{Setup}
\myparagraph{Dataset Construction.}
\label{sec:dataset_construction}
We built a subject-augmented image-text paired dataset based on the FFHQ-wild~\cite{karras2019style} dataset to train our models.
First, we use the BLIP-2 model~\cite{li2023blip} \texttt{blip2-opt-6.7b-coco} to generate captions for all images. Next, we employ the Mask2Former model~\cite{cheng2022masked} \texttt{mask2former-swin-large-coco-panoptic} to generate panoptic segmentation masks for each image. 
We then leverage the spaCy~\cite{spacy2} library to chunk all noun phrases in the image captions and expand numbered plural phrases (\eg, "two women") into singular phrases connected by ``and'' (\eg, "a woman and a woman"). 
Finally, we use a greedy matching algorithm to match noun phrases with image segments. We do this by considering the product of the image-text similarity score by the OpenCLIP model~\cite{ilharco_gabriel_2021_5143773} \texttt{CLIP-ViT-H-14-laion2B-s32B-b79K} and the label-text similarity score by the Sentence-Transformer~\cite{reimers2019sentence} model \texttt{stsb-mpnet-base-v2}.
We reserve 1000 images for validation and testing purposes.

\myparagraph{Training Details.}
We start training from the StableDiffusion v1-5~\cite{rombach2022high} model. 
To encode the visual inputs, we use OpenAI's \texttt{clip-vit-large-patch14} vision model, which serves as the partner model of the text encoder in SDv1-5. 
During training, we freeze the text encoder and only train the U-Net, the MLP module, and the last two transformer blocks of the vision encoder. 
We train our models for 150k steps on 8 NVIDIA A6000 GPUs, with a constant learning rate of 1e-5 and a batch size of 128. 
We only augment segments whose COCO~\cite{lin2014coco} label is ``person'' and set a maximum of 4 reference subjects during training, with each subject having a 10\% chance of being dropped.
We train the model solely on text conditioning with 10\% of the samples to maintain the model's capability for text-only generation. To facilitate classifier-free guidance sampling~\cite{ho2022classifier}, we train the model without any conditions on 10\% of the instances.
During training, we apply the loss only in the subject region to half of the training samples to enhance the generation quality in the subject area.

\myparagraph{Evaluation Metric}

We evaluate image generation quality on identity preservation and prompt consistency.
Identity preservation is determined by detecting faces in the reference and generated images using MTCNN~\cite{zhang2016joint}, and then calculating a pairwise identity similarity using FaceNet~\cite{schroff2015facenet}. 
For multi-subject evaluation, we identify all faces within the generated images and use a greedy matching procedure between the generated faces and reference subjects. 
The \emph{minimum} similarity value among all subjects measures overall identity preservation.
We evaluate the prompt consistency using the average CLIP-L/14 image-text similarity following textual-inversion~\cite{gal2022image}.
For efficiency evaluation, we consider the total time for customization, including fine-tuning (for tuning-based methods) and inference. 
We also measure peak memory usage during the entire procedure.
\subsection{Single-Subject Image Generation}
\label{sec:single}
Our first evaluation targets the performance of single-subject image generation.
Given the lack of published baselines in our tuning-free environment, we compare with leading optimization-based approaches, including DreamBooth~\cite{ruiz2022dreambooth}, Textual-Inversion~\cite{gal2022image}, and Custom Diffusion~\cite{kumari2022multi}.
We use the implementations from diffusers library~\cite{diffusers}. 
We provide the detailed hyperparameters in the appendix section. 
We assess the capabilities of these different methods in generating personalized content for subjects derived from the Celeb-A dataset~\cite{liu2015deep}. 
To construct our evaluation benchmark, we develop a broad range of text prompts encapsulating a wide spectrum of scenarios, such as recontextualization, stylization, accessorization, and diverse actions. 
The entire test set comprises 15 subjects, with 30 unique text prompts allocated to each. An exhaustive list of text prompts is available in the appendix.
We utilized five images per subject to fine-tune the optimization-based methods, given our observation that these methods overfit and simply reproduce the reference image when a single reference image is used. 
In contrast, our model employs a single randomly selected image for each subject.
Shown in Table~\ref{tab:single_subject}, \method surpasses all baselines, delivering superior identity preservation and prompt consistency. 
Remarkably, it achieves 300$\times$-1200$\times$ speedup and 2.8$\times$ reduction in memory usage. 
Figure \ref{fig:single_subject} shows the qualitative results of single-subject personalization comparisons, employing different approaches across an array of prompts.
Significantly, our model matches the text consistency of text-only methods and exceeds all baseline strategies in terms of identity preservation, with only single input and forward passes used.
\begin{table}[t]
    \centering
    \small
    \caption{Comparison between our method and baseline approaches on single-subject image generation. StableDiffusion served as the text-only baseline without any subject conditioning.}
    \label{tab:single_subject}

    \begin{tabular}{l@{\ \ }c@{\ \ }c@{\ \ }c@{\ \ }c@{\ \ }c}
        \toprule
        Method                & Images $\downarrow$ & Identity Preservation $\uparrow$ & Prompt Consistency $\uparrow$ & Total Time $\downarrow$ & Peak Memory $\downarrow$ \\
        \midrule
        StableDiffusion       & 0                   & 3.85\%                 & 26.79\%             & 2s                      & 6 GB                     \\
        \midrule
        Textual-Inversion     & 5                   & 29.26\%               & 21.91\%             & 2500 s                  & 17 GB                    \\
        DreamBooth            & 5                   & 27.27\%              & 23.91\%             & 1084 s                  & 40 GB                    \\
        Custom Diffusion      & 5                   & 43.37\%               & 23.29\%             & 789 s                   & 29 GB                    \\

        \textbf{FastComposer} & \textbf{1}          & \textbf{51.41\%}      & \textbf{24.30\%}    & \textbf{2 s}            & \textbf{6 GB}            \\
        \bottomrule
    \end{tabular}
    \vspace{-1em}
\end{table}

\subsection{Multi-Subject Image Generation}
We then consider a more complex setting: multi-object, subject-driven image generation. We examine the quality of multi-subject generation by using all possible combinations (105 pairs in total) formed from 15 subjects described in Sec.~\ref{sec:single}, allocating 21 prompts to each pair for assessment.
Table~\ref{tab:multi_subject} shows a quantitative analysis contrasting \method with the baseline methods. 
Optimization-based methods~\cite{kumari2022multi, ruiz2022dreambooth, gal2022image} frequently falter in maintaining identity preservation, often generating generic images or images that blend identities among different reference subjects.
\method, on the other hand, preserves the unique features of different subjects, yielding a significantly improved identity preservation score.
Furthermore, our prompt consistency is on par with tuning-based approaches~\cite{gal2022image, kumari2022multi}. 
Qualitative comparisons are shown in Figure~\ref{fig:multi_subject}.
More visual examples for three-subject images are shown in Figure~\ref{fig:subject3}.
\begin{table}[t]
    \centering
    \small
    \caption{Comparison between our method and baseline approaches on multiple-subject Image generation. StableDiffusion served as the text-only baseline without any subject conditioning.}
    \label{tab:multi_subject}
    \begin{tabular}{l@{\ \ }c@{\ \ }c@{\ \ }c@{\ \ }c@{\ \ }c}
        \toprule
        Method                & Images $\downarrow$ & Identity Preservation $\uparrow$ & Prompt Consistency $\uparrow$ & Total Time $\downarrow$ & Peak Memory $\downarrow$ \\
        \midrule
        StableDiffusion       & 0                   & 1.88\%                 & 28.44\%             & 2 s                     & 6 GB                     \\
        \midrule
        Textual-Inversion     & 5                   & 13.52\%               & 21.08\%             & 4998  s                 & 17 GB                    \\
        Custom Diffusion      & 5                   & 5.37\%                            & \textbf{25.84\%}                & 789 s                   & 29 GB                    \\
        \textbf{FastComposer} & \textbf{1}          & \textbf{43.11\%}      & 24.25\%                        & \textbf{2 s}            & \textbf{6 GB}            \\
        \bottomrule
    \end{tabular}
    \vspace{-1em}
\end{table}

\begin{minipage}{\textwidth}
  \begin{minipage}[t]{0.44\textwidth}\vspace{0pt}
    \centering
    \includegraphics[width=\linewidth]{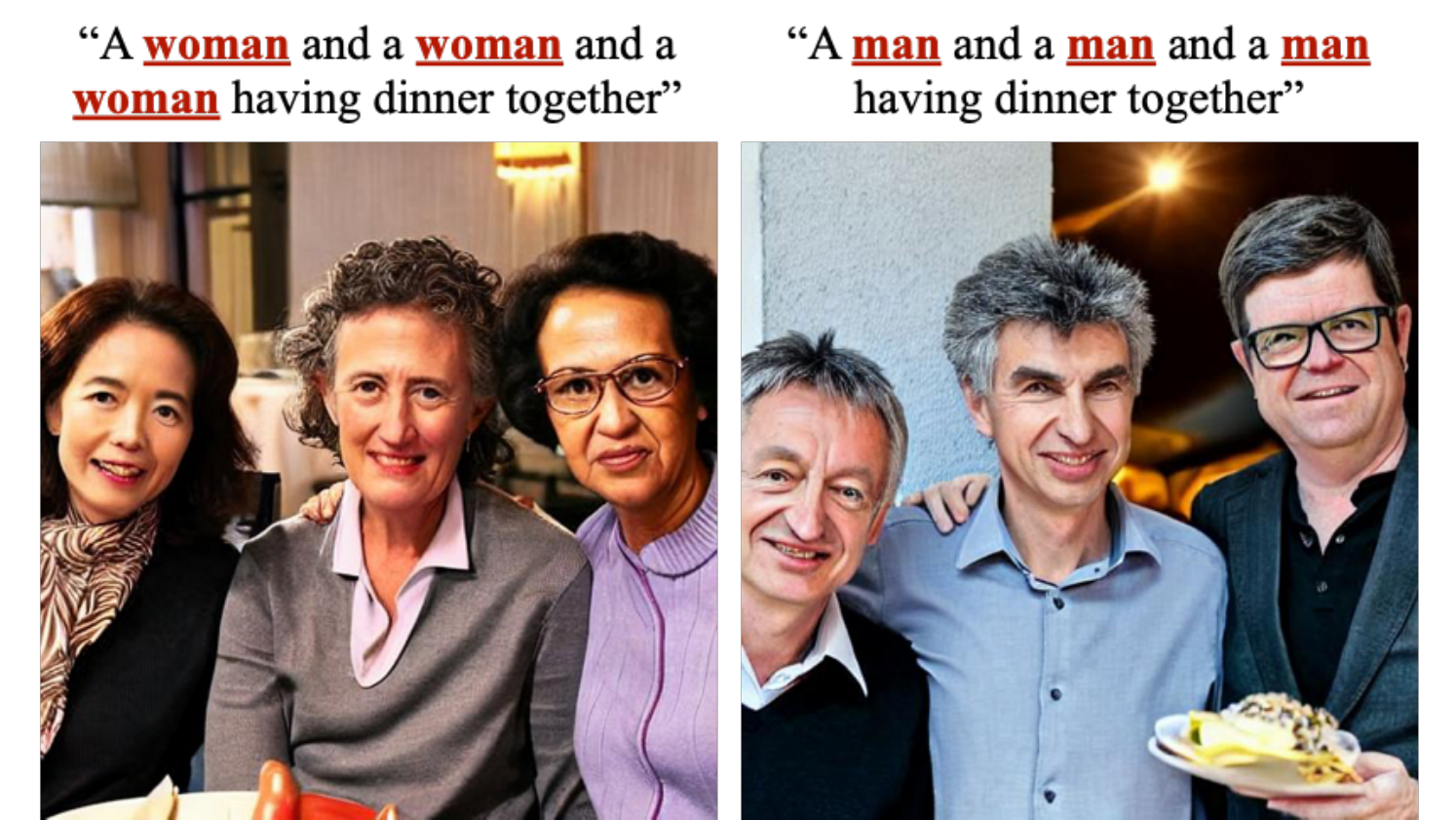}
    \captionof{figure}{Generating images with three subjects.}
    \label{fig:subject3}
  \end{minipage}
  \hfill
  \begin{minipage}[t]{0.55\textwidth}\vspace{0pt}
    \centering
    \small
\setlength{\tabcolsep}{3.0pt}
      \captionof{table}{Ablation studies on the cross-attention localization supervision. We compare with the model trained in the same setting without cross-attention localization.}
    \label{tab:abalation_loc}
    \begin{tabular}{lccc}
        \toprule
        Method         & Identity Pres. $\uparrow$ & Prompt Cons. $\uparrow$ & \\
        \midrule
        w/o Loc.       & 37.66\%              & \textbf{25.03\%}             \\
        w/ Loc. (Ours) & \textbf{43.11\%}      & 24.25\%      \\
        \bottomrule
    \end{tabular}

    \end{minipage}
  \end{minipage}

\subsection{Ablation Study}
\myparagraph{Delayed Subject Conditioning.}
Figure \ref{fig:timesteps} shows the impact of varying the ratio of timesteps devoted to subject conditioning, a hyperparameter in our \textit{delayed subject conditioning} approach. 
As this ratio increases, the model improves in identity preservation but loses editability.
A ratio between 0.6 to 0.8 achieves a favorable balance on the tradeoff curve. 

\myparagraph{Cross-Attention Localization Loss.}
Table \ref{tab:abalation_loc} presents the ablation studies on our proposed \textit{cross-attention localization} loss. 
The baseline is trained in the same setting but excludes the localization loss. 
Our method demonstrates a substantial enhancement of the identity preservation score. 
Figure~\ref{fig:crossattn} shows the qualitative comparisons.
Incorporating the localization loss allows the model to focus on particular reference subjects, thereby avoiding identity blending.

\section{Discussion and Conclusion}

We propose \method, a tuning-free method for personalized, multi-subject text-to-image generation. We achieve tuning-free subject-driven image generation by using a pre-trained vision encoder, making this process efficient and accessible across various platforms.
\method effectively tackles the identity blending issue in multi-subject generation by supervising cross-attention maps with segmentation masks during training. 
We also propose a novel delayed subject conditioning technique to balance the identity preservation and the flexibility of image editability. 

\paragraph{Limitations.} 
First, the current training set is FFHQ~\cite{karras2019style} which is small and primarily contains headshots of human faces. 
It also has a long-tailed distribution for the number of people, thus limiting our ability to generate images with more than three subjects.   
Utilizing a more diverse dataset will enable \method to generate a broader range of actions and scenarios, thereby enhancing its versatility and applicability.
Second, our work is primarily human-centric due to a scarcity of large-scale, multi-subject datasets featuring other subjects like animals. 
We believe that broadening our dataset to incorporate multi-subject imagery of other categories will significantly enrich our model's capabilities.
Finally, our model, built on the foundation of Stable Diffusion and FFHQ, also inherits their biases.

\section*{Acknowledgements}
This work is supported by 
MIT AI Hardware Program,
NVIDIA Academic Partnership Award,
MIT-IBM Watson AI Lab,  
Amazon and MIT Science Hub,   
Microsoft Turing Academic Program,
Singapore DSTA under DST00OECI20300823 (New Representations for Vision),
NSF grant 2105819 and NSF CAREER Award 1943349.
\newpage
\small
\bibliographystyle{plain}
\bibliography{reference}

\begin{thebibliography}{10}

\bibitem{avrahami2022spatext}
Omri Avrahami, Thomas Hayes, Oran Gafni, Sonal Gupta, Yaniv Taigman, Devi
  Parikh, Dani Lischinski, Ohad Fried, and Xi~Yin.
\newblock Spatext: Spatio-textual representation for controllable image
  generation.
\newblock {\em CVPR}, 2023.

\bibitem{bolya2022token}
Daniel Bolya, Cheng-Yang Fu, Xiaoliang Dai, Peizhao Zhang, Christoph
  Feichtenhofer, and Judy Hoffman.
\newblock Token merging: Your {ViT} but faster.
\newblock In {\em International Conference on Learning Representations}, 2023.

\bibitem{casanova2021instance}
Arantxa Casanova, Marlene Careil, Jakob Verbeek, Michal Drozdzal, and Adriana
  Romero~Soriano.
\newblock Instance-conditioned gan.
\newblock {\em Advances in Neural Information Processing Systems},
  34:27517--27529, 2021.

\bibitem{chang2023muse}
Huiwen Chang, Han Zhang, Jarred Barber, AJ~Maschinot, Jose Lezama, Lu~Jiang,
  Ming-Hsuan Yang, Kevin Murphy, William~T Freeman, Michael Rubinstein, et~al.
\newblock Muse: Text-to-image generation via masked generative transformers.
\newblock {\em arXiv preprint arXiv:2301.00704}, 2023.

\bibitem{chefer2023attend}
Hila Chefer, Yuval Alaluf, Yael Vinker, Lior Wolf, and Daniel Cohen-Or.
\newblock Attend-and-excite: Attention-based semantic guidance for
  text-to-image diffusion models.
\newblock {\em Siggraph}, 2023.

\bibitem{chen2016training}
Tianqi Chen, Bing Xu, Chiyuan Zhang, and Carlos Guestrin.
\newblock Training deep nets with sublinear memory cost, 2016.

\bibitem{cheng2022masked}
Bowen Cheng, Ishan Misra, Alexander~G Schwing, Alexander Kirillov, and Rohit
  Girdhar.
\newblock Masked-attention mask transformer for universal image segmentation.
\newblock In {\em Proceedings of the IEEE/CVF Conference on Computer Vision and
  Pattern Recognition}, pages 1290--1299, 2022.

\bibitem{ding2021cogview}
Ming Ding, Zhuoyi Yang, Wenyi Hong, Wendi Zheng, Chang Zhou, Da~Yin, Junyang
  Lin, Xu~Zou, Zhou Shao, Hongxia Yang, et~al.
\newblock Cogview: Mastering text-to-image generation via transformers.
\newblock {\em Advances in Neural Information Processing Systems},
  34:19822--19835, 2021.

\bibitem{gal2022image}
Rinon Gal, Yuval Alaluf, Yuval Atzmon, Or~Patashnik, Amit~H Bermano, Gal
  Chechik, and Daniel Cohen-Or.
\newblock An image is worth one word: Personalizing text-to-image generation
  using textual inversion.
\newblock {\em ICLR}, 2023.

\bibitem{gal2023designing}
Rinon Gal, Moab Arar, Yuval Atzmon, Amit~H Bermano, Gal Chechik, and Daniel
  Cohen-Or.
\newblock Designing an encoder for fast personalization of text-to-image
  models.
\newblock {\em Siggraph}, 2023.

\bibitem{han2015deep}
Song Han, Huizi Mao, and William~J Dally.
\newblock Deep compression: Compressing deep neural networks with pruning,
  trained quantization and huffman coding.
\newblock {\em ICLR}, 2016.

\bibitem{hertz2022prompt}
Amir Hertz, Ron Mokady, Jay Tenenbaum, Kfir Aberman, Yael Pritch, and Daniel
  Cohen-Or.
\newblock Prompt-to-prompt image editing with cross attention control.
\newblock {\em ICLR}, 2023.

\bibitem{ho2020denoising}
Jonathan Ho, Ajay Jain, and Pieter Abbeel.
\newblock Denoising diffusion probabilistic models.
\newblock {\em Advances in Neural Information Processing Systems},
  33:6840--6851, 2020.

\bibitem{ho2022classifier}
Jonathan Ho and Tim Salimans.
\newblock Classifier-free diffusion guidance.
\newblock {\em arXiv preprint arXiv:2207.12598}, 2022.

\bibitem{spacy2}
Matthew Honnibal and Ines Montani.
\newblock {spaCy 2}: Natural language understanding with {B}loom embeddings,
  convolutional neural networks and incremental parsing.
\newblock To appear, 2017.

\bibitem{ilharco_gabriel_2021_5143773}
Gabriel Ilharco, Mitchell Wortsman, Ross Wightman, Cade Gordon, Nicholas
  Carlini, Rohan Taori, Achal Dave, Vaishaal Shankar, Hongseok Namkoong, John
  Miller, Hannaneh Hajishirzi, Ali Farhadi, and Ludwig Schmidt.
\newblock Openclip, July 2021.

\bibitem{kang2023scaling}
Minguk Kang, Jun-Yan Zhu, Richard Zhang, Jaesik Park, Eli Shechtman, Sylvain
  Paris, and Taesung Park.
\newblock Scaling up gans for text-to-image synthesis.
\newblock {\em CVPR}, 2023.

\bibitem{karras2019style}
Tero Karras, Samuli Laine, and Timo Aila.
\newblock A style-based generator architecture for generative adversarial
  networks.
\newblock In {\em Proceedings of the IEEE/CVF conference on computer vision and
  pattern recognition}, pages 4401--4410, 2019.

\bibitem{kirstain2023x}
Yuval Kirstain, Omer Levy, and Adam Polyak.
\newblock X\&fuse: Fusing visual information in text-to-image generation.
\newblock {\em arXiv preprint arXiv:2303.01000}, 2023.

\bibitem{kumari2022multi}
Nupur Kumari, Bingliang Zhang, Richard Zhang, Eli Shechtman, and Jun-Yan Zhu.
\newblock Multi-concept customization of text-to-image diffusion.
\newblock {\em CVPR}, 2023.

\bibitem{li2023blip}
Junnan Li, Dongxu Li, Silvio Savarese, and Steven Hoi.
\newblock Blip-2: Bootstrapping language-image pre-training with frozen image
  encoders and large language models.
\newblock {\em arXiv preprint arXiv:2301.12597}, 2023.

\bibitem{lin2014coco}
Tsung-Yi Lin, Michael Maire, Serge Belongie, Lubomir Bourdev, Ross Girshick,
  James Hays, Pietro Perona, Deva Ramanan, C.~Lawrence Zitnick, and Piotr
  Dollár.
\newblock Microsoft coco: Common objects in context, 2014.

\bibitem{liu2015deep}
Ziwei Liu, Ping Luo, Xiaogang Wang, and Xiaoou Tang.
\newblock Deep learning face attributes in the wild.
\newblock In {\em Proceedings of the IEEE international conference on computer
  vision}, pages 3730--3738, 2015.

\bibitem{ma2023unified}
Yiyang Ma, Huan Yang, Wenjing Wang, Jianlong Fu, and Jiaying Liu.
\newblock Unified multi-modal latent diffusion for joint subject and text
  conditional image generation.
\newblock {\em arXiv preprint arXiv:2303.09319}, 2023.

\bibitem{nitzan2022mystyle}
Yotam Nitzan, Kfir Aberman, Qiurui He, Orly Liba, Michal Yarom, Yossi
  Gandelsman, Inbar Mosseri, Yael Pritch, and Daniel Cohen-Or.
\newblock Mystyle: A personalized generative prior.
\newblock {\em ACM Transactions on Graphics (TOG)}, 41(6):1--10, 2022.

\bibitem{radford2021learning}
Alec Radford, Jong~Wook Kim, Chris Hallacy, Aditya Ramesh, Gabriel Goh,
  Sandhini Agarwal, Girish Sastry, Amanda Askell, Pamela Mishkin, Jack Clark,
  et~al.
\newblock Learning transferable visual models from natural language
  supervision.
\newblock In {\em International conference on machine learning}, pages
  8748--8763. PMLR, 2021.

\bibitem{ramesh2022hierarchical}
Aditya Ramesh, Prafulla Dhariwal, Alex Nichol, Casey Chu, and Mark Chen.
\newblock Hierarchical text-conditional image generation with clip latents.
\newblock {\em arXiv preprint arXiv:2204.06125}, 2022.

\bibitem{ramesh2021zero}
Aditya Ramesh, Mikhail Pavlov, Gabriel Goh, Scott Gray, Chelsea Voss, Alec
  Radford, Mark Chen, and Ilya Sutskever.
\newblock Zero-shot text-to-image generation.
\newblock In {\em International Conference on Machine Learning}, pages
  8821--8831. PMLR, 2021.

\bibitem{reimers2019sentence}
Nils Reimers and Iryna Gurevych.
\newblock Sentence-bert: Sentence embeddings using siamese bert-networks.
\newblock {\em EMNLP}, 2019.

\bibitem{roich2022pivotal}
Daniel Roich, Ron Mokady, Amit~H Bermano, and Daniel Cohen-Or.
\newblock Pivotal tuning for latent-based editing of real images.
\newblock {\em ACM Transactions on Graphics (TOG)}, 42(1):1--13, 2022.

\bibitem{rombach2022high}
Robin Rombach, Andreas Blattmann, Dominik Lorenz, Patrick Esser, and Bj{\"o}rn
  Ommer.
\newblock High-resolution image synthesis with latent diffusion models.
\newblock In {\em Proceedings of the IEEE/CVF Conference on Computer Vision and
  Pattern Recognition}, pages 10684--10695, 2022.

\bibitem{ruiz2022dreambooth}
Nataniel Ruiz, Yuanzhen Li, Varun Jampani, Yael Pritch, Michael Rubinstein, and
  Kfir Aberman.
\newblock Dreambooth: Fine tuning text-to-image diffusion models for
  subject-driven generation.
\newblock {\em CVPR}, 2023.

\bibitem{sarukkai2023collage}
Vishnu Sarukkai, Linden Li, Arden Ma, Christopher R{\'e}, and Kayvon
  Fatahalian.
\newblock Collage diffusion.
\newblock {\em arXiv preprint arXiv:2303.00262}, 2023.

\bibitem{schroff2015facenet}
Florian Schroff, Dmitry Kalenichenko, and James Philbin.
\newblock Facenet: A unified embedding for face recognition and clustering.
\newblock In {\em Proceedings of the IEEE conference on computer vision and
  pattern recognition}, pages 815--823, 2015.

\bibitem{shi2023instantbooth}
Jing Shi, Wei Xiong, Zhe Lin, and Hyun~Joon Jung.
\newblock Instantbooth: Personalized text-to-image generation without test-time
  finetuning.
\newblock {\em arXiv preprint arXiv:2304.03411}, 2023.

\bibitem{sohl2015deep}
Jascha Sohl-Dickstein, Eric Weiss, Niru Maheswaranathan, and Surya Ganguli.
\newblock Deep unsupervised learning using nonequilibrium thermodynamics.
\newblock In {\em International Conference on Machine Learning}, pages
  2256--2265. PMLR, 2015.

\bibitem{song2020score}
Yang Song, Jascha Sohl-Dickstein, Diederik~P Kingma, Abhishek Kumar, Stefano
  Ermon, and Ben Poole.
\newblock Score-based generative modeling through stochastic differential
  equations.
\newblock {\em ICLR}, 2021.

\bibitem{diffusers}
Patrick von Platen, Suraj Patil, Anton Lozhkov, Pedro Cuenca, Nathan Lambert,
  Kashif Rasul, Mishig Davaadorj, and Thomas Wolf.
\newblock Diffusers: State-of-the-art diffusion models.
\newblock \url{https://github.com/huggingface/diffusers}, 2022.

\bibitem{wei2023elite}
Yuxiang Wei, Yabo Zhang, Zhilong Ji, Jinfeng Bai, Lei Zhang, and Wangmeng Zuo.
\newblock Elite: Encoding visual concepts into textual embeddings for
  customized text-to-image generation.
\newblock {\em arXiv preprint arXiv:2302.13848}, 2023.

\bibitem{xiao2022smoothquant}
Guangxuan Xiao, Ji~Lin, Mickael Seznec, Julien Demouth, and Song Han.
\newblock Smoothquant: Accurate and efficient post-training quantization for
  large language models.
\newblock {\em arXiv preprint arXiv:2211.10438}, 2022.

\bibitem{zhang2016joint}
Kaipeng Zhang, Zhanpeng Zhang, Zhifeng Li, and Yu~Qiao.
\newblock Joint face detection and alignment using multitask cascaded
  convolutional networks.
\newblock {\em IEEE signal processing letters}, 23(10):1499--1503, 2016.

\end{thebibliography}

\end{document}